\pdfoutput=1
\documentclass[11pt]{article}

\usepackage[preprint]{acl}
\usepackage{times}
\usepackage{latexsym}
\usepackage[T1]{fontenc}
\usepackage[utf8]{inputenc}
\usepackage{microtype}
\usepackage{inconsolata}
\usepackage{graphicx}

\usepackage{amsmath} 
\usepackage{bbm}

\usepackage{enumitem}

\usepackage{forest}
\usepackage{xcolor}

\usepackage{url}

\title{Prompt Compression for Large Language Models: A Survey}

\author{Zongqian Li\thanks{Co-first authors.} \and Yinhong Liu\footnotemark[1] \and Yixuan Su \and Nigel Collier \\
    University of Cambridge \\
    \texttt{\{zl510, yl535, ys484, nhc30\}@cam.ac.uk}}
         
\begin{document}
\maketitle
\begin{abstract}
Leveraging large language models (LLMs) for complex natural language tasks typically requires long-form prompts to convey detailed requirements and information, which results in increased memory usage and inference costs. To mitigate these challenges, multiple efficient methods have been proposed, with prompt compression gaining significant research interest. This survey provides an overview of prompt compression techniques, categorized into hard prompt methods and soft prompt methods. First, the technical approaches of these methods are compared, followed by an exploration of various ways to understand their mechanisms, including the perspectives of attention optimization, Parameter-Efficient Fine-Tuning (PEFT), modality integration, and new synthetic language. We also examine the downstream adaptations of various prompt compression techniques. Finally, the limitations of current prompt compression methods are analyzed, and several future directions are outlined, such as optimizing the compression encoder, combining hard and soft prompts methods, and leveraging insights from multimodality.
\end{abstract}

\section{Introduction}

\begin{figure}[th!]
    \centering
    \includegraphics[width=0.49\textwidth]{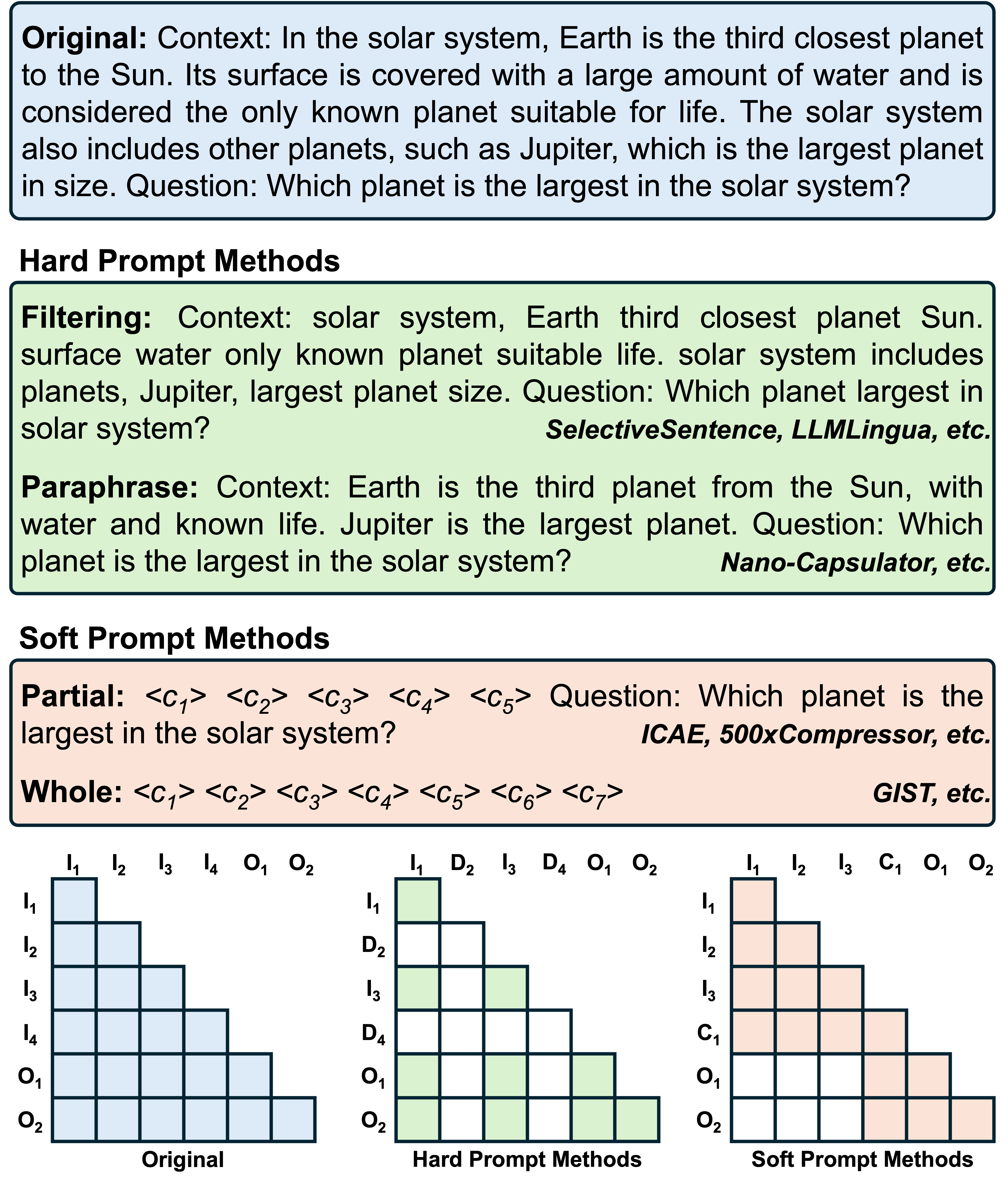}
    \caption{Illustrative examples of prompt compression methods. Hard prompt methods remove low-information tokens or paraphrase for conciseness. Soft prompt methods compress text into a smaller number of special tokens, $<c_n>$. The grids below visualize attention patterns, where the y-axis represents the sequence of tokens, and the x-axis shows the tokens they attend to. In the original prompt, each token attends to all previous tokens. In hard prompts, each token cannot attend to previous deleted tokens ($D_i$). In soft prompts, after the compressed token ($C_i$) attends to all prior input tokens ($I_i$), subsequent output tokens ($O_i$) cannot attend to tokens before the compressed token.}
    \label{cover_figure}
\end{figure}

As task complexity increases, prompts become longer for LLMs to accommodate more detailed requirements, contextual information, and in-context learning (ICL) examples. Lengthy prompts reduce inference speed, increase memory costs, and negatively impact user experience. Current methods for improving LLM efficiency can be broadly categorized into model-centric and prompt-centric methods \citep{wan2024efficient}. Model-centric approaches, such as parameter pruning \citep{NEURIPS2023_44956951} and quantization \citep{NEURIPS2022_c3ba4962}, focus on optimizing the model itself. In contrast, prompt-centric methods, including prompt compression \citep{li-etal-2023-compressing} and prompt design \citep{shin-etal-2020-autoprompt}, aim to improve the efficiency of LLM by lowering the complexity of input. Prompt-centric methods typically introduce minimal or no changes to the parameters of the LLM, allowing them to be used in a plug-and-play fashion. Thus, these methods, especially for prompt compression as shown in Figure~\ref{cover_figure}, have gained increasing attention. However, the optimal architectures and underlying mechanisms of prompt compression remain unclear, highlighting an important area for further investigation.

\begin{figure*}[th!]
    \centering
    \includegraphics[width=0.97\textwidth]{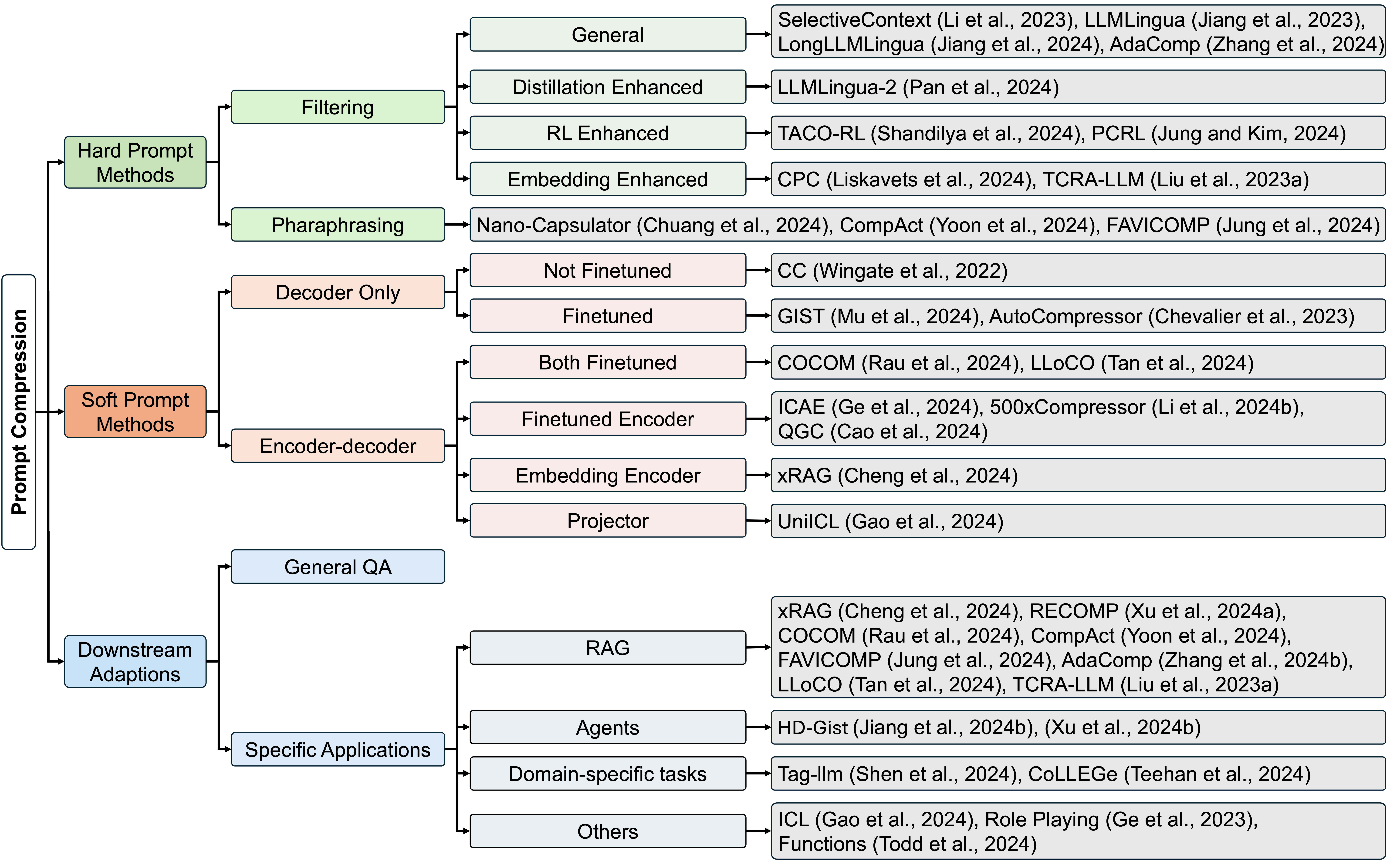}
    \caption{Hierarchical overview of prompt compression methods and their downstream adaptions.}
    \label{tree_overview}
\end{figure*}

This survey aims to introduce LLM prompt compression to researchers with prior knowledge of attention mechanisms, transformers, and LLMs. Unlike previous surveys on general efficient prompting \citep{chang2024efficientpromptingmethodslarge} or prompting for reasoning \citep{qiao-etal-2023-reasoning}, this paper specifically focuses on prompt compression. The preliminary knowledge is introduced in Section~\ref{Preliminary}, covering prompting and efficiency. This is followed by a detailed discussion of hard prompt methods in Section~\ref{Hard Prompt Methods} and soft prompt methods in Section~\ref{Soft Prompt Methods}. The key factors driving the evolution of prompt compression models are examined, and insights into various methods are provided. Their downstream adaptations are discussed in Section~\ref{Downstream Adaptions} as well. Finally, Section~\ref{Challenges and Future Work} analyzes the challenges of current prompt compression techniques and proposes several potential solutions. An overview of prompt compression methods and their downstream adaptations is shown in Figure~\ref{tree_overview}.

This survey highlights that:\vspace{-0.7\baselineskip}

\begin{itemize}[left=0pt, itemsep=0pt, parsep=0pt]
    \item In addition to the exploration of existing prompt compression methods, we provide our interpretation of their mechanisms: filtering or paraphrasing in hard prompt methods, and attention modification, PEFT, modality integration, or new synthetic language in soft prompt methods.
    \item We identify challenges in current methods, including fine-tuning problems such as overfitting and performance degradation, high compression latency, and the need for comparison with attention optimization techniques.
    \item We propose several future research directions to solve current challenges, including optimizing the compression encoder, combining hard and soft prompts, and integrating insights from multimodal LLMs.
\end{itemize}

\section{Preliminary}
\label{Preliminary}

\textbf{Prompts} are important for LLMs, broadly classified into hard and soft prompts \citep{10.1145/3560815}. They serve as input instructions to guide LLMs in performing tasks such as summarization, classification, translation, and question answering without the need for fine-tuning \citep{vatsal2024surveypromptengineeringmethods}. The flexibility and effectiveness of prompts leverage the generalization abilities of LLMs, influenced by factors such as wording, in-context examples, clarity, and accuracy, making prompt design a critical area of research \citep{schulhoff2024promptreportsystematicsurvey}.

\textbf{Prompt structures} in this survey consist of two main components: an instruction or question and an input or context, depending on the task. The {instruction, input, output} format is commonly used in instruction fine-tuning datasets such as Alpaca \citep{taori2023alpaca} and PwC \citep{ge2024incontext} for tasks such as content creation and machine translation. The {context, question, answer} format is prevalent in QA and reading comprehension datasets such as SQuAD \citep{rajpurkar-etal-2016-squad}, HotpotQA \citep{yang-etal-2018-hotpotqa}, and RACE \citep{lai-etal-2017-race}. Prompts can vary significantly in length, with different prompt compression methods targeting some or all of these components.

\textbf{Hard prompts} are natural language prompts made up of tokens from the vocabulary set of the LLM, corresponding to specific words or sub-words \citep{sennrich-etal-2016-neural}. These prompts can be generated by either humans or models. While natural language prompts are interpretable and provide transparency, their inherent ambiguity can make it difficult to fully capture intent in a concise manner. This limitation reduces the utility of hard prompts in diverse contexts or complex scenarios. Additionally, creating effective and precise hard prompts requires considerable human effort, and can involve training a model to refine or optimize these prompts. Furthermore, variations in hard prompts can lead to differences in LLM performance for the same task.

\textbf{Soft prompts} are trainable, continuous vectors that share the same dimensions as token embeddings in the dictionary of the LLM \citep{zhao-etal-2023-spc}. Different from hard prompts, the tokens in soft prompts are higly trained to convey nuanced meanings that cannot be captured by discrete tokens in the predefined vocabulary. When tuned on diverse datasets, soft prompts are expected to help the LLM perform well across different tasks. However, as the dataset size grows, the computational resources needed increase as well. Additionally, soft prompts are less explainable by humans compared to hard prompts, as their tokens are not directly readable.

\textbf{Prompt compression} aims to reduce the length of prompts, thereby improving the efficiency of processing LLM inputs \citep{wan2024efficient}. There are two primary approaches to prompt compression: removing unnecessary or low-information content (hard prompt methods) and learning continuous representations of the prompt information in the embedding space (soft prompt methods) \citep{chang2024efficientpromptingmethodslarge}. Hard prompt methods act as a form of filtering, still using natural language tokens, although the resulting prompts may be less fluent and grammatically correct, and can potentially generalize to LLMs with different embedding configurations. Soft prompt methods, on the other hand, employ encoding to convert prompt information into continuous representations, resulting in latent vectors (special tokens) that cannot be understood by humans.

\section{Hard Prompt Methods}
\label{Hard Prompt Methods}

Hard prompt methods remove unnecessary tokens from the original prompt while maintaining the use of natural language words or sub-words \citep{10.1145/3411763.3451760}. These methods are particularly useful for LLMs that only accept natural language inputs, such as black-box API models, rather than word embeddings. There are three representative hard prompt methods: SelectiveContext \citep{li-etal-2023-compressing} and LLMLingua \citep{jiang-etal-2023-llmlingua} for filtering, and Nano-Capsulator \citep{chuang-etal-2024-learning} for paraphrasing, as shown in Figure~\ref{hard_prompt_methods}.

\textbf{SelectiveContext} identifies and deletes redundant or less informative parts of an input prompt by quantifying the informativeness of lexical units using self-information \citep{li-etal-2023-compressing}. To maintain text coherence, the syntactic parsing capabilities of Spacy\footnote{\url{https://spacy.io/api/pipeline-functions\#merge_noun_chunks}} are used to group individual tokens into noun phrases based on dependency parsing. SelectiveContext does not rely on external models or additional parameters, and can be applied to any model architecture. However, there are three main disadvantages: (1) The filtered prompts must be re-encoded by the LLM, limiting potential efficiency gains. (2) It relies on accurate phrase boundary detection using SpaCy. (3) Currently, there are no methods for merging verb phrases.

\textbf{LLMLingua} employs a smaller language model, such as GPT-2 \citep{radford2019language}, to calculate the self-information or perplexity of content, similar to SelectiveContext, and removes redundant tokens from the natural language prompt before it is fed to the LLM \citep{jiang-etal-2023-llmlingua}. LLMLingua operates on prompts structured as \{Instruction, Input (Demonstrations), Question\}, initially selecting key demonstrations based on perplexity scores. It then applies token-level filtering across the prompt, allowing for breaking words into sub-word units and avoiding noun phrase merging. For key elements such as numbers and units, LLMLingua incorporates token preservation algorithms that prioritize these elements within instructions and questions. Achieving compression ratios up to 20x, the compression process of LLMLingua is managed by the smaller external language model, allowing it to work with black-box LLMs. However, there are two limitations: (1) The smaller language model requires additional memory and may use a different tokenizer than the larger LLM. (2) Since not all prompts contain a large portion of in-context examples (demonstrations), it is important to differentiate between prompt compression and in-context example selection.

\textbf{Nano-Capsulator} summarizes the original prompt into a concise natural language version, which is then input into the LLM \citep{chuang-etal-2024-learning}. This process removes irrelevant information and reconstructs the prompt into fluent sentences. The compression model, fine-tuned Vicuna-7B, operates independently of the LLM. Different from standard summarization, Nano-Capsulator includes a semantic preservation loss to retain key meanings important for downstream tasks and a reward function to optimize the utility of the prompt for the LLM. This targeted approach yields better task performance by enhancing semantic relevance and compatibility. However, the memory cost of the compression model is not negligible. Additionally, the compression process is akin to a pre-generation step, requiring more computational resources due to the need for additional inference rather than simple encoding.

In addition to the methods mentioned above, general hard prompt methods include LongLLMLingua, which has a longer compression window than LLMLingua by applying document reordering and subsequence recovery \citep{jiang-etal-2024-longllmlingua}, and AdaComp, an adaptive compression method that dynamically selects relevant documents based on query complexity and retrieval quality \citep{zhang2024adacomp}. These general methods have been further improved in different ways. For example, LLMLingua-2 uses data distillation to create a compressed dataset and trains a classifier to retain essential tokens \citep{pan-etal-2024-llmlingua}. Both PCRL \citep{Jung_2024} and TACO-RL \citep{shandilya2024TACO-RL} apply reinforcement learning (RL) for token selection: PCRL is model-agnostic, while TACO-RL is task-specific. CPC \citep{liskavets2024CPC} and TCRA-LLM \citep{liu-etal-2023-tcra} reduce tokens by leveraging embeddings: CPC ranks sentence relevance with context-aware embeddings, while TCRA-LLM uses embeddings for summarization and semantic compression.

\section{Soft Prompt Methods}
\label{Soft Prompt Methods}

\subsection{Architectures}

Soft prompt compression methods typically consist of two main components: an encoder that compresses the prompts into a shorter sequence of continuous special tokens and a decoder that processes the compressed prompts to generate corresponding responses. As these models have evolved, they demonstrate improved generalization abilities, require fewer additional trainable parameters in the original LLMs, allow for longer prompt lengths to be compressed, and achieve higher compression ratios. Several common architectures for these compression models are illustrated in Figure~\ref{soft_prompt_methods}, including contrastive conditioning (CC) \citep{wingate-etal-2022-prompt}, gist tokens (GIST) \citep{mu2024learning}, AutoCompressor \citep{chevalier2023adapting}, in-context autoencoder (ICAE) \citep{ge2024incontext}, 500xCompressor \citep{500xcompressor}, xRAG \citep{cheng2024xragextremecontextcompression}, and UniICL \citep{gao2024unifyingdemonstrationselectioncompression}.

\textbf{CC} is a decoder-only method that trains a shorter soft prompt to approximate the output distribution of a natural language prompt by minimizing the Kullback-Leibler (KL) divergence across token sequences, thereby aligning with desired response patterns \citep{wingate-etal-2022-prompt}. The output distributions are estimated through repeated sequence sampling, with both the natural and soft prompts serving as conditional inputs for token generation. By applying various contrastive contexts, soft prompts can be trained to produce specific attributes, such as enhanced sentiment, allowing for effective content control. However, each soft prompt is uniquely trained for a specific natural language prompt, which limits the generalization capabilities of CC, as new prompts require retraining from scratch.

\textbf{GIST} modifies the attention mechanism of the LLM \citep{mu2024learning}. Compressed tokens, a series of a new trainable token, are appended after the original prompt tokens. While these tokens can attend to the original prompt, newly generated tokens can only attend to the compressed tokens, enforcing a separation in attention flow. This setup can be viewed as an encoder-decoder architecture: an LLM fine-tuned on Alpaca+ functions as the encoder, compressing the original prompt into a smaller number of compressed tokens. These tokens then serve as input to the decoder (the same fine-tuned LLM), which generates corresponding responses. Different from CC, GIST generalizes to compress unseen prompts without additional fine-tuning. Although GIST achieves a compression ratio of up to 26x, it is limited by the maximum compressive prompt length, due to the short prompts in the fine-tuning dataset. Additionally, the compressed tokens cannot be used with the original untuned LLM, restricting its broader applicability.

\textbf{AutoCompressor} shares a similar architecture with GIST and can handle long context prompt compression up to 30,720 tokens \citep{chevalier2023adapting}. The whole process is recursive, with the original prompt divided into several sub-prompts. In each iteration, the sub-prompt is compressed into a small set of tokens, which are then passed to the next iteration along with a new sub-prompt for further compression. This continues until all the compressed tokens are collected, representing the information from the entire original prompt. While AutoCompressor increases the maximum prompt length that can be compressed, the training process is time-consuming, and the compressed tokens cannot be used by the original untuned LLM.

\textbf{ICAE} increases the compression length and uses the frozen LLM as the decoder \citep{ge2024incontext}. It compresses long, information-rich contexts into a small number of tokens, which are then used for QA. Unlike GIST, which compresses low-information texts of around 30 tokens focused on questions and instructions, ICAE can handle detailed, complex contexts. The question itself remains uncompressed, and the answer is generated based on the compressed context and the uncompressed question. Since the decoder is frozen, the compressed tokens can be used directly with the original LLM without fine-tuning. ICAE compresses up to 512 tokens into 32, 64, or 128 special tokens, achieving compression ratios between 4x and 16x. By concatenating multiple groups of compressed tokens, ICAE can handle up to 5,120 tokens. However, its compression ratio decreases compared to GIST. In addition, ICAE is trained and tested on the Pile dataset \citep{gao2020pile}, which may overlap with the training corpus of the LLM, raising concerns about potential data leakage and the possibility of retrieving answers from the memory of the LLM.

\textbf{500xCompressor} explores prompt compression at high compression ratios while maintaining the compression length limit of ICAE \citep{500xcompressor}. Similar to ICAE, 500xCompressor employs trainable LoRA parameters \citep{gao2020pile800gbdatasetdiverse} in the encoder, while keeping the original LLM frozen in the decoder. However, different from ICAE, 500xCompressor feeds the decoder with the K V values of the compressed tokens rather than the tokens themselves. These K V values keep more detailed information than embedding vectors, especially under high compression ratios. 500xCompressor uses 1, 4, or 16 tokens to compress 96, 192, 288, 384, or 480 tokens, achieving compression ratios from 6x to 480x, while retaining 62.26\% to 72.89\% of the capabilities of the uncompressed prompts. Moreover, the test set ArxivQA, constructed from arXiv abstracts from January to April 2024, ensures evaluation on strictly unseen data, mitigating potential data leakage risks.

\textbf{xRAG} uses a frozen embedding model as the encoder, with only an adapter, positioned between the encoder and the decoder LLM, containing trainable parameters. \citep{cheng2024xragextremecontextcompression}. Although primarily designed for Retrieval-Augmented Generation (RAG) \citep{NEURIPS2020_6b493230}, the task of xRAG remains a QA task, making it applicable as a general prompt compression method. xRAG proves that current embedding models can compress information into a single token for QA tasks. While several embedding models were tested in the original paper, the final choice was SFR-Embedding-Mistral \citep{SFRAIResearch2024}, which is still based on the LLM and requires substantial memory. As a result, xRAG needs to load two LLMs and a projector, whereas ICAE and 500xCompressor only need to load a single LLM and a set of LoRA parameters.

\textbf{UniICL} focuses on compressing the demonstrations component of the input prompt, leaving the query unchanged \citep{gao2024unifyingdemonstrationselectioncompression}. The only trainable component in UniICL is a projector placed between the encoder and decoder. Unlike previous soft prompt methods, both the encoder and decoder in UniICL are frozen and utilize the same LLM, reducing gradient computation during training and conserving memory for loading the LLM. In ICL, the quality and relevance of the examples influence model performance. The compressed tokens in UniICL can be considered as embeddings for various in-context examples, eliminating the need for additional embedding processes during in-context example selection.

Besides the seven methods introduced above, several other similar approaches exist, such as COCOM \citep{rau2024cocom}, LLoCO \citep{tan2024LLoCO}, and QGC \citep{cao-etal-2024-retaining}. COCOM and LLoCO use fine-tuned encoder-decoder setups, while QGC employs a frozen decoder. Both COCOM and LLoCO are designed for the task of RAG: COCOM compresses multiple documents into groups of context embeddings and inputs them together into the decoder, while LLoCO stores and retrieves specific LoRA parameters for the decoder to adapt to particular text types and tasks.

\subsection{Insights}
\label{Insights}

Soft prompt methods for prompt compression can be understood from several perspectives, including attention mechanism optimization, analogies with prompt and prefix tuning, encoding natural language into a new modality, and viewing compressed prompts as a new language for LLMs. 

\textbf{Attention mechanism.} In the standard self-attention of transformer, each newly generated token attends to all previous tokens, which increases computational costs along with the sequence length. Soft prompt compression reduces the input length in two stages: first, a small number of special tokens attend to the full input, storing key information in these tokens; second, new tokens are generated based solely on the compressed tokens. This process blocks the need for the full input during generation, limiting attention to the compressed tokens and reducing computation. This can be seen as a form of attention optimization, though it differs in that the K V values generated in the first stage are not identical to those in the original LLM (as LoRA parameters or adapters are added). There are other attention optimization methods as well, including sliding window attention \citep{Beltagy2020Longformer} and sparse attention \citep{child2019generating}. In sliding window attention, for example, all input tokens are kept, however, each token can only attend to a limited number of preceding tokens.

\textbf{Prompt and prefix tuning.} Prompt tuning \citep{lester-etal-2021-power} and prefix tuning \citep{li-liang-2021-prefix} are PEFT methods for language models. In prompt tuning, a set of trainable embeddings is added to the beginning of the input. In prefix tuning, both the added embeddings and their corresponding K V values in each layer of the language model are trainable as well. While these methods are useful for fine-tuning models for specific tasks, their significance has diminished with the advent of LLMs capable of handling downstream tasks through simple prompting. ICAE and related methods are similar to prompt tuning. Instead of manually training the embeddings, these embeddings are generated by the encoder (a fine-tuned LLM) through compressing the natural language input. Each natural language prompt is encoded into a unique set of embeddings, with different prompts corresponding to different sets. This process provides a zero-shot way of determining the parameters typically trained in prompt tuning. In contrast, 500xCompressor is more akin to prefix tuning. Here, the inputs for the decoder are not embeddings, but K V values, which are determined by the fine-tuned LLM encoder. These K V values contain more parameters than embeddings and are assumed to store richer details, however, they increase the risk of overfitting to specific tasks \citep{li-liang-2021-prefix}. Nonetheless, 500xCompressor proves that K V values can compress more detailed information at high compression ratios and outperform embeddings in terms of data retention \citep{500xcompressor}. Since both embeddings and K V values are generated simultaneously by the encoder, they offer similar improvements in LLM efficiency, including higher inference speed and reduced computational costs.

\textbf{Modality integration.} Compressed texts can be considered a new modality, similar to vision and speech. In vision-language models, for example, an image encoder converts an image into a list of embeddings, which are then used by an LLM decoder for downstream tasks such as image captioning or QA \citep{10445007}. This architecture parallels that of prompt compression models, where an encoder compresses natural language into special embeddings and an LLM decoder utilizes these compressed tokens. The training processes for both are similar as well: image encoders are trained to predict missing parts of an image and align with textual descriptions, while prompt compression encoders are trained to regenerate compressed texts and interact with natural language inputs. Thus, compressed tokens from text can be seen as a new, rich-information modality derived from natural language. However, it is important to note that texts contain more specific details and have a higher information density than images. As a result, text compression requires greater precision than image encoding, and the potential for information loss during prompt compression is more significant than during image encoding.

\textbf{New synthetic language.} A new language can be defined by three key features: (1) it can encode information and convert thoughts, concepts, or data into representations; (2) this encoded information can be transmitted between entities, whether individuals or systems; and (3) the system can dynamically adjust its interpretation or output based on the input it receives, a process known as adaptive evaluation \citep{mansilla2004foundations}. In the context of prompt compression, the compressed tokens represent natural language texts as vectors in latent space, which can be understood by LLMs. These compressed tokens can be saved, transferred between different LLMs, and facilitate knowledge transfer. Moreover, when the encoded information changes, LLMs can dynamically adjust their responses based on the new input. Therefore, the compressed tokens generated by prompt compression models can be considered a new, more efficient language for LLMs. Research is ongoing in the definition and evaluation of this emerging LLM-specific language \citep{guo2020inductive}.

\section{Downstream Adaptions}
\label{Downstream Adaptions}

Prompt compression has a wide range of applications in the domains such as general QA, RAG \citep{cheng2024xragextremecontextcompression, xu2024recomp, rau2024cocom, yoon2024compact, jung2024FaviComp, zhang2024adacomp, tan2024LLoCO, liu-etal-2023-tcra}, ICL \citep{gao2024unifyingdemonstrationselectioncompression}, role playing \citep{NEURIPS2023_6fcbfb37}, agent-based \citep{jiang-etal-2024-hierarchical, xu2024conciseprecisecontextcompression}, and interdisciplinary tasks \citep{shen2024tag, teehan2024college}. In general QA, prompt compression methods are generally used to compress instructions \citep{mu2024learning} or contexts \citep{500xcompressor} to perform various instruction-following tasks \citep{ge2024incontext}. In RAG, xRAG exemplifies a method that answers questions based on text embeddings encoded from retrieved documents rather than processing the entire text \citep{cheng2024xragextremecontextcompression}. For ICL, UniICL is an example method that compresses in-context examples to a smaller number of embedding tokens, helping with the selection of relevant examples \citep{gao2024unifyingdemonstrationselectioncompression}. In agent systems, API documentation can be compressed to enable more efficient tool use \citep{jiang-etal-2024-hierarchical}.

\section{Challenges and Future Work}
\label{Challenges and Future Work}

\subsection{Current Challenges}

Current prompt compression methods face several challenges, including information loss, reduced model capability, and marginal improvements in efficiency, due to fine-tuning limitations and inefficient compression processes.

\textbf{Finetuning problems.} Although some soft prompt methods, such as ICAE, 500xCompressor, xRAG, and UniICL, do not need to fine-tune the decoder LLM, the input soft prompt functions similarly to prompt tuning or prefix tuning for the decoder, leading to problems related to fine-tuning. Previous research has shown that fine-tuning base models can result in catastrophic forgetting, overfitting, and model drift, which can degrade the generalization performance of the base LLM \citep{gururangan-etal-2020-dont}. To avoid these problems, prompt compression models must be trained on large, diverse datasets, which is computationally expensive. Furthermore, the trained encoders are specific to the corresponding decoder LLMs, meaning that when the LLM is updated, for example, from LLaMA-2 \cite{touvron2023llama} to LLaMA-3 \cite{dubey2024llama3herdmodels}, the encoder must be re-trained. Hard prompts face challenges as well: filtered hard prompts may disrupt grammatical correctness and provide an unfamiliar input distribution to the LLM, potentially affecting its performance.

\textbf{Limited efficiency improvements.} Although prompt length can be reduced by dozens or even hundreds of times, the time required for compression and the memory needed to store the compressor remain under optimized. Current encoders in soft prompt methods can be finetuned using either full-parameter training or PEFT methods, such as LoRA and adapters. Fully fine-tuned LLMs are the most resource-intensive, while the additional parameters in LoRA and adapter-based encoders are plug-and-play, allowing for separate loading. However, larger encoders result in longer compression times. Hardware variability can lead to fluctuations in time for the same task as well \citep{hennessy2011computer}, introducing large standard deviations and relative errors, making it essential to perform multiple evaluations for accuracy. Moreover, if the encoder and decoder are of equal size in soft prompt methods, the computations required for prompt compression are nearly the same or even slightly higher than those for encoding the original prompt. As a result, efficiency gains are only realized during the generation of new tokens. For tasks with short outputs, these improvements may not be substantial. Hard prompt methods also face issues: additional models may be needed to determine which tokens to delete, and the filtered prompts still need to be re-encoded by the LLM, further impacting efficiency.

\textbf{Comparison with attention modification methods.} As discussed in Section~\ref{Insights}, soft prompt methods can be regarded as specialized attention modifications. However, current prompt compression methods have not been compared to traditional attention optimization methods, such as sliding window attention \citep{Beltagy2020Longformer} and sparse attention \citep{child2019generating}. Unlike soft prompt methods, attention modifications do not need an encoder model, which eliminates additional memory costs \citep{NEURIPS2021_bd4a6d05}. Additionally, in soft prompt methods, the input and generated tokens rely on different attention mechanisms, whereas attention optimization methods apply the same mechanism to both the input and the generated tokens, resulting in greater stability and scalability \citep{10.1145/3530811}. Therefore, it is important to determine the compression ratio when both methods have equivalent computational requirements and compare prompt compression methods with attention modification methods.

\subsection{Future Directions}

To address current challenges and enhance prompt compression methods, several future directions are proposed to reduce information loss while increasing compression ratio and speed.

\textbf{Encoder Optimization.} In current soft prompt methods, encoders are similar in size to decoders. For instance, if a decoder has 8 billion parameters, the encoder may add tens of millions of trainable parameters on top of that. As a result, the time used for compression is comparable to the time it takes for the original LLM to process inputs, meaning that soft prompt methods only improve efficiency during the generation of new tokens. In theory, large encoders that are similar to the decoders can compress information well from the original text into compressed tokens for the LLM. However, smaller, well-trained models such as BERT, which have fewer parameters (as least 10 times smaller than LLMs), are capable of encoding semantic information effectively \citep{devlin-etal-2019-bert}. This will substantially increase compression speed. Besides LoRA and adapters, other PEFT methods, such as QLoRA \citep{NEURIPS2023_1feb8787}, DoRA \citep{liu2024dora}, MoRA \citep{jiang2024mora}, and LLaMA-Adapter \citep{zhang2024llamaadapter} are worth trying for fine-tuning the compression encoder as well.

\textbf{Combination of hard and soft prompts.} Hard prompt methods increase information density by filtering out unnecessary tokens. Soft prompt methods represent the original text using a small number of special tokens. Since hard and soft prompt methods operate through orthogonal mechanisms, their combination can further enhance compression ratios. However, one challenge is that the compression times for hard and soft prompts cannot be parallelized, which can reduce the overall efficiency of encoding LLM inputs.

\textbf{Insights from multimodal LLMs.} As discussed in Section~\ref{Insights}, soft prompt methods can be understood as a form of modality integration between natural language and compressed language. This opens the possibility of applying insights from multimodal LLMs to benefit prompt compression models. There are mainly two ways to encode images into embeddings: self-attention and cross-attention \citep{jin2024efficientmultimodallargelanguage}. In self-attention, images and query vectors are input to the image encoder together, whereas in cross-attention, only query vectors are input to the encoder, and they attend to external image embeddings in each layer. Current soft prompt methods rely on self-attention to transfer information from natural language prompts to compressed tokens, however, cross-attention remains unexplored. Trying other multimodal architectures, such as those using cross-attention, may offer new ways to leverage compressed tokens. In image-text representation integration, image encoders are trained to align image embeddings with natural language embeddings, a process that can be adapted for prompt compression models \citep{pmlr-v139-radford21a}.

\section{Conclusions}

This survey provides a comprehensive overview of the previous prompt compression methods, from the perspectives of hard and soft prompt approaches. In addition to discussing the technical approaches of these models, different perspectives on understanding the compression process and their applications are explored. Furthermore, we also discuss the challenges of the existing prompt compression methods and suggest the potential future development directions. It should be noted that prompt compression methods should be compared with attention optimization techniques, and they can benefit from insights drawn from multimodal LLMs. We hope our survey offers a comprehensive overview of existing methods, providing deeper insights into their motivations and technical approaches. We also aim for our suggested research directions to inspire the community and support future advancements in the field.

\section*{Limitations}
This paper focuses specifically on prompt compression and does not provide an overview of all efficient prompting methods or other efficiency-related LLM techniques. Rather than addressing a broad topic, it offers a detailed explanation and insights into the specific area of prompt compression. It should be noted that the use of any prompt compression methods should follow their guidelines and copyright restrictions.

\section*{Ethics Statement}
This research did not involve any studies with human participants or animals performed by any of the authors. Therefore, no ethical approval was required for this study. All data and materials were collected in a manner consistent with ethical guidelines, ensuring no ethical concerns are present.

\section*{Availability Statement}
The list of papers and updates related to this survey are uploaded to the open-source community at https://github.com/ZongqianLi/Prompt-Compression-Survey.

\section*{Acknowledgement}
Thanks Zheng Hui for proofreading this article.

\bibliography{custom}

\begin{thebibliography}{70}
\providecommand{\natexlab}[1]{#1}

\bibitem[{Beltagy et~al.(2020)Beltagy, Peters, and Cohan}]{Beltagy2020Longformer}
Iz~Beltagy, Matthew~E. Peters, and Arman Cohan. 2020.
\newblock Longformer: The long-document transformer.
\newblock \emph{arXiv:2004.05150}.

\bibitem[{Cao et~al.(2024)Cao, Cao, Lu, Peng, Huang, Cheng, and Su}]{cao-etal-2024-retaining}
Zhiwei Cao, Qian Cao, Yu~Lu, Ningxin Peng, Luyang Huang, Shanbo Cheng, and Jinsong Su. 2024.
\newblock \href {https://doi.org/10.18653/v1/2024.acl-long.685} {Retaining key information under high compression ratios: Query-guided compressor for {LLM}s}.
\newblock In \emph{Proceedings of the 62nd Annual Meeting of the Association for Computational Linguistics (Volume 1: Long Papers)}, pages 12685--12695, Bangkok, Thailand. Association for Computational Linguistics.

\bibitem[{Chang et~al.(2024)Chang, Xu, Wang, Luo, Xiao, and Zhu}]{chang2024efficientpromptingmethodslarge}
Kaiyan Chang, Songcheng Xu, Chenglong Wang, Yingfeng Luo, Tong Xiao, and Jingbo Zhu. 2024.
\newblock \href {https://arxiv.org/abs/2404.01077} {Efficient prompting methods for large language models: A survey}.
\newblock \emph{Preprint}, arXiv:2404.01077.

\bibitem[{Cheng et~al.(2024)Cheng, Wang, Zhang, Ge, Chen, Wei, Zhang, and Zhao}]{cheng2024xragextremecontextcompression}
Xin Cheng, Xun Wang, Xingxing Zhang, Tao Ge, Si-Qing Chen, Furu Wei, Huishuai Zhang, and Dongyan Zhao. 2024.
\newblock \href {https://arxiv.org/abs/2405.13792} {xrag: Extreme context compression for retrieval-augmented generation with one token}.
\newblock \emph{Preprint}, arXiv:2405.13792.

\bibitem[{Chevalier et~al.(2023)Chevalier, Wettig, Ajith, and Chen}]{chevalier2023adapting}
Alexis Chevalier, Alexander Wettig, Anirudh Ajith, and Danqi Chen. 2023.
\newblock \href {https://openreview.net/forum?id=kp1U6wBPXq} {Adapting language models to compress contexts}.
\newblock In \emph{The 2023 Conference on Empirical Methods in Natural Language Processing}.

\bibitem[{Child et~al.(2019)Child, Gray, Radford, and Sutskever}]{child2019generating}
Rewon Child, Scott Gray, Alec Radford, and Ilya Sutskever. 2019.
\newblock Generating long sequences with sparse transformers.
\newblock \emph{arXiv preprint arXiv:1904.10509}.

\bibitem[{Chuang et~al.(2024)Chuang, Xing, Chang, Liu, Chen, and Hu}]{chuang-etal-2024-learning}
Yu-Neng Chuang, Tianwei Xing, Chia-Yuan Chang, Zirui Liu, Xun Chen, and Xia Hu. 2024.
\newblock \href {https://doi.org/10.18653/v1/2024.naacl-long.429} {Learning to compress prompt in natural language formats}.
\newblock In \emph{Proceedings of the 2024 Conference of the North American Chapter of the Association for Computational Linguistics: Human Language Technologies (Volume 1: Long Papers)}, pages 7756--7767, Mexico City, Mexico. Association for Computational Linguistics.

\bibitem[{Dettmers et~al.(2022)Dettmers, Lewis, Belkada, and Zettlemoyer}]{NEURIPS2022_c3ba4962}
Tim Dettmers, Mike Lewis, Younes Belkada, and Luke Zettlemoyer. 2022.
\newblock \href {https://proceedings.neurips.cc/paper_files/paper/2022/file/c3ba4962c05c49636d4c6206a97e9c8a-Paper-Conference.pdf} {Gpt3.int8(): 8-bit matrix multiplication for transformers at scale}.
\newblock In \emph{Advances in Neural Information Processing Systems}, volume~35, pages 30318--30332. Curran Associates, Inc.

\bibitem[{Dettmers et~al.(2023)Dettmers, Pagnoni, Holtzman, and Zettlemoyer}]{NEURIPS2023_1feb8787}
Tim Dettmers, Artidoro Pagnoni, Ari Holtzman, and Luke Zettlemoyer. 2023.
\newblock \href {https://proceedings.neurips.cc/paper_files/paper/2023/file/1feb87871436031bdc0f2beaa62a049b-Paper-Conference.pdf} {Qlora: Efficient finetuning of quantized llms}.
\newblock In \emph{Advances in Neural Information Processing Systems}, volume~36, pages 10088--10115. Curran Associates, Inc.

\bibitem[{Devlin et~al.(2019)Devlin, Chang, Lee, and Toutanova}]{devlin-etal-2019-bert}
Jacob Devlin, Ming-Wei Chang, Kenton Lee, and Kristina Toutanova. 2019.
\newblock \href {https://doi.org/10.18653/v1/N19-1423} {{BERT}: Pre-training of deep bidirectional transformers for language understanding}.
\newblock In \emph{Proceedings of the 2019 Conference of the North {A}merican Chapter of the Association for Computational Linguistics: Human Language Technologies, Volume 1 (Long and Short Papers)}, pages 4171--4186, Minneapolis, Minnesota. Association for Computational Linguistics.

\bibitem[{Dubey et~al.(2024)Dubey, Jauhri, Pandey, Kadian, Al-Dahle, ..., and Zhao}]{dubey2024llama3herdmodels}
Abhimanyu Dubey, Abhinav Jauhri, Abhinav Pandey, Abhishek Kadian, Ahmad Al-Dahle, ..., and Zhiwei Zhao. 2024.
\newblock \href {https://arxiv.org/abs/2407.21783} {The llama 3 herd of models}.
\newblock \emph{Preprint}, arXiv:2407.21783.

\bibitem[{Gao et~al.(2024)Gao, Cao, and Li}]{gao2024unifyingdemonstrationselectioncompression}
Jun Gao, Ziqiang Cao, and Wenjie Li. 2024.
\newblock \href {https://arxiv.org/abs/2405.17062} {Unifying demonstration selection and compression for in-context learning}.
\newblock \emph{Preprint}, arXiv:2405.17062.

\bibitem[{Gao et~al.(2020{\natexlab{a}})Gao, Biderman, Black, Golding, Hoppe, Foster, Phang, He, Thite, Nabeshima, Presser, and Leahy}]{gao2020pile800gbdatasetdiverse}
Leo Gao, Stella Biderman, Sid Black, Laurence Golding, Travis Hoppe, Charles Foster, Jason Phang, Horace He, Anish Thite, Noa Nabeshima, Shawn Presser, and Connor Leahy. 2020{\natexlab{a}}.
\newblock \href {https://arxiv.org/abs/2101.00027} {The pile: An 800gb dataset of diverse text for language modeling}.
\newblock \emph{Preprint}, arXiv:2101.00027.

\bibitem[{Gao et~al.(2020{\natexlab{b}})Gao, Biderman, Black, Golding, Hoppe, Foster, Phang, He, Thite, Nabeshima et~al.}]{gao2020pile}
Leo Gao, Stella Biderman, Sid Black, Laurence Golding, Travis Hoppe, Charles Foster, Jason Phang, Horace He, Anish Thite, Noa Nabeshima, et~al. 2020{\natexlab{b}}.
\newblock The pile: An 800gb dataset of diverse text for language modeling.
\newblock \emph{arXiv preprint arXiv:2101.00027}.

\bibitem[{Ge et~al.(2023)Ge, Jing, Dong, Mao, Xia, Wang, Chen, and Wei}]{NEURIPS2023_6fcbfb37}
Tao Ge, Hu~Jing, Li~Dong, Shaoguang Mao, Yan Xia, Xun Wang, Si-Qing Chen, and Furu Wei. 2023.
\newblock \href {https://proceedings.neurips.cc/paper_files/paper/2023/file/6fcbfb3721c1781728b10c6685cc2f6c-Paper-Conference.pdf} {Extensible prompts for language models on zero-shot language style customization}.
\newblock In \emph{Advances in Neural Information Processing Systems}, volume~36, pages 35576--35591. Curran Associates, Inc.

\bibitem[{Ge et~al.(2024)Ge, Jing, Wang, Wang, Chen, and Wei}]{ge2024incontext}
Tao Ge, Hu~Jing, Lei Wang, Xun Wang, Si-Qing Chen, and Furu Wei. 2024.
\newblock \href {https://openreview.net/forum?id=uREj4ZuGJE} {In-context autoencoder for context compression in a large language model}.
\newblock In \emph{The Twelfth International Conference on Learning Representations}.

\bibitem[{Guo et~al.(2020)Guo, Ren, S{\l}owik, and Mathewson}]{guo2020inductive}
Shangmin Guo, Yi~Ren, Agnieszka S{\l}owik, and Kory Mathewson. 2020.
\newblock Inductive bias and language expressivity in emergent communication.
\newblock \emph{4th NeurIPS Workshop on Emergent Communication}.

\bibitem[{Gururangan et~al.(2020)Gururangan, Marasovi{\'c}, Swayamdipta, Lo, Beltagy, Downey, and Smith}]{gururangan-etal-2020-dont}
Suchin Gururangan, Ana Marasovi{\'c}, Swabha Swayamdipta, Kyle Lo, Iz~Beltagy, Doug Downey, and Noah~A. Smith. 2020.
\newblock \href {https://doi.org/10.18653/v1/2020.acl-main.740} {Don{'}t stop pretraining: Adapt language models to domains and tasks}.
\newblock In \emph{Proceedings of the 58th Annual Meeting of the Association for Computational Linguistics}, pages 8342--8360, Online. Association for Computational Linguistics.

\bibitem[{Hennessy and Patterson(2011)}]{hennessy2011computer}
John~L Hennessy and David~A Patterson. 2011.
\newblock \emph{Computer architecture: a quantitative approach}.
\newblock Elsevier.

\bibitem[{Jiang et~al.(2023)Jiang, Wu, Lin, Yang, and Qiu}]{jiang-etal-2023-llmlingua}
Huiqiang Jiang, Qianhui Wu, Chin-Yew Lin, Yuqing Yang, and Lili Qiu. 2023.
\newblock \href {https://doi.org/10.18653/v1/2023.emnlp-main.825} {{LLML}ingua: Compressing prompts for accelerated inference of large language models}.
\newblock In \emph{Proceedings of the 2023 Conference on Empirical Methods in Natural Language Processing}, pages 13358--13376, Singapore. Association for Computational Linguistics.

\bibitem[{Jiang et~al.(2024{\natexlab{a}})Jiang, Wu, Luo, Li, Lin, Yang, and Qiu}]{jiang-etal-2024-longllmlingua}
Huiqiang Jiang, Qianhui Wu, Xufang Luo, Dongsheng Li, Chin-Yew Lin, Yuqing Yang, and Lili Qiu. 2024{\natexlab{a}}.
\newblock \href {https://aclanthology.org/2024.acl-long.91} {{L}ong{LLML}ingua: Accelerating and enhancing {LLM}s in long context scenarios via prompt compression}.
\newblock In \emph{Proceedings of the 62nd Annual Meeting of the Association for Computational Linguistics (Volume 1: Long Papers)}, pages 1658--1677, Bangkok, Thailand. Association for Computational Linguistics.

\bibitem[{Jiang et~al.(2024{\natexlab{b}})Jiang, Huang, Luo, Zhang, Huang, Wei, Deng, Sun, Zhang, Wang et~al.}]{jiang2024mora}
Ting Jiang, Shaohan Huang, Shengyue Luo, Zihan Zhang, Haizhen Huang, Furu Wei, Weiwei Deng, Feng Sun, Qi~Zhang, Deqing Wang, et~al. 2024{\natexlab{b}}.
\newblock Mora: High-rank updating for parameter-efficient fine-tuning.
\newblock \emph{arXiv preprint arXiv:2405.12130}.

\bibitem[{Jiang et~al.(2024{\natexlab{c}})Jiang, Vecchio, Bansal, and Johannsen}]{jiang-etal-2024-hierarchical}
Yichen Jiang, Marco Vecchio, Mohit Bansal, and Anders Johannsen. 2024{\natexlab{c}}.
\newblock \href {https://aclanthology.org/2024.findings-eacl.143} {Hierarchical and dynamic prompt compression for efficient zero-shot {API} usage}.
\newblock In \emph{Findings of the Association for Computational Linguistics: EACL 2024}, pages 2162--2174, St. Julian{'}s, Malta. Association for Computational Linguistics.

\bibitem[{Jin et~al.(2024)Jin, Li, Liu, Gu, Wu, Jiang, He, Zhao, Tan, Gan, Wang, Wang, and Ma}]{jin2024efficientmultimodallargelanguage}
Yizhang Jin, Jian Li, Yexin Liu, Tianjun Gu, Kai Wu, Zhengkai Jiang, Muyang He, Bo~Zhao, Xin Tan, Zhenye Gan, Yabiao Wang, Chengjie Wang, and Lizhuang Ma. 2024.
\newblock \href {https://arxiv.org/abs/2405.10739} {Efficient multimodal large language models: A survey}.
\newblock \emph{Preprint}, arXiv:2405.10739.

\bibitem[{Jung et~al.(2024)Jung, Liu, Huang, Zhou, and Chen}]{jung2024FaviComp}
Dongwon Jung, Qin Liu, Tenghao Huang, Ben Zhou, and Muhao Chen. 2024.
\newblock \href {https://arxiv.org/abs/2409.12468} {Familiarity-aware evidence compression for retrieval augmented generation}.
\newblock \emph{Preprint}, arXiv:2409.12468.

\bibitem[{Jung and Kim(2024)}]{Jung_2024}
Hoyoun Jung and Kyung-Joong Kim. 2024.
\newblock \href {https://doi.org/10.1109/access.2024.3403426} {Discrete prompt compression with reinforcement learning}.
\newblock \emph{IEEE Access}, 12:72578–72587.

\bibitem[{Lai et~al.(2017)Lai, Xie, Liu, Yang, and Hovy}]{lai-etal-2017-race}
Guokun Lai, Qizhe Xie, Hanxiao Liu, Yiming Yang, and Eduard Hovy. 2017.
\newblock \href {https://doi.org/10.18653/v1/D17-1082} {{RACE}: Large-scale {R}e{A}ding comprehension dataset from examinations}.
\newblock In \emph{Proceedings of the 2017 Conference on Empirical Methods in Natural Language Processing}, pages 785--794, Copenhagen, Denmark. Association for Computational Linguistics.

\bibitem[{Lester et~al.(2021)Lester, Al-Rfou, and Constant}]{lester-etal-2021-power}
Brian Lester, Rami Al-Rfou, and Noah Constant. 2021.
\newblock \href {https://doi.org/10.18653/v1/2021.emnlp-main.243} {The power of scale for parameter-efficient prompt tuning}.
\newblock In \emph{Proceedings of the 2021 Conference on Empirical Methods in Natural Language Processing}, pages 3045--3059, Online and Punta Cana, Dominican Republic. Association for Computational Linguistics.

\bibitem[{Lewis et~al.(2020)Lewis, Perez, Piktus, Petroni, Karpukhin, Goyal, K\"{u}ttler, Lewis, Yih, Rockt\"{a}schel, Riedel, and Kiela}]{NEURIPS2020_6b493230}
Patrick Lewis, Ethan Perez, Aleksandra Piktus, Fabio Petroni, Vladimir Karpukhin, Naman Goyal, Heinrich K\"{u}ttler, Mike Lewis, Wen-tau Yih, Tim Rockt\"{a}schel, Sebastian Riedel, and Douwe Kiela. 2020.
\newblock \href {https://proceedings.neurips.cc/paper_files/paper/2020/file/6b493230205f780e1bc26945df7481e5-Paper.pdf} {Retrieval-augmented generation for knowledge-intensive nlp tasks}.
\newblock In \emph{Advances in Neural Information Processing Systems}, volume~33, pages 9459--9474. Curran Associates, Inc.

\bibitem[{Li and Liang(2021)}]{li-liang-2021-prefix}
Xiang~Lisa Li and Percy Liang. 2021.
\newblock \href {https://doi.org/10.18653/v1/2021.acl-long.353} {Prefix-tuning: Optimizing continuous prompts for generation}.
\newblock In \emph{Proceedings of the 59th Annual Meeting of the Association for Computational Linguistics and the 11th International Joint Conference on Natural Language Processing (Volume 1: Long Papers)}, pages 4582--4597, Online. Association for Computational Linguistics.

\bibitem[{Li et~al.(2023)Li, Dong, Guerin, and Lin}]{li-etal-2023-compressing}
Yucheng Li, Bo~Dong, Frank Guerin, and Chenghua Lin. 2023.
\newblock \href {https://doi.org/10.18653/v1/2023.emnlp-main.391} {Compressing context to enhance inference efficiency of large language models}.
\newblock In \emph{Proceedings of the 2023 Conference on Empirical Methods in Natural Language Processing}, pages 6342--6353, Singapore. Association for Computational Linguistics.

\bibitem[{Li et~al.(2024)Li, Su, and Collier}]{500xcompressor}
Zongqian Li, Yixuan Su, and Nigel Collier. 2024.
\newblock \href {https://arxiv.org/abs/2408.03094} {500xcompressor: Generalized prompt compression for large language models}.
\newblock \emph{Preprint}, arXiv:2408.03094.

\bibitem[{Liskavets et~al.(2024)Liskavets, Ushakov, Roy, Klibanov, Etemad, and Luke}]{liskavets2024CPC}
Barys Liskavets, Maxim Ushakov, Shuvendu Roy, Mark Klibanov, Ali Etemad, and Shane Luke. 2024.
\newblock \href {https://arxiv.org/abs/2409.01227} {Prompt compression with context-aware sentence encoding for fast and improved llm inference}.
\newblock \emph{Preprint}, arXiv:2409.01227.

\bibitem[{Liu et~al.(2023{\natexlab{a}})Liu, Li, Xiang, Wang, and Qian}]{liu-etal-2023-tcra}
Junyi Liu, Liangzhi Li, Tong Xiang, Bowen Wang, and Yiming Qian. 2023{\natexlab{a}}.
\newblock \href {https://doi.org/10.18653/v1/2023.findings-emnlp.655} {{TCRA}-{LLM}: Token compression retrieval augmented large language model for inference cost reduction}.
\newblock In \emph{Findings of the Association for Computational Linguistics: EMNLP 2023}, pages 9796--9810, Singapore. Association for Computational Linguistics.

\bibitem[{Liu et~al.(2023{\natexlab{b}})Liu, Yuan, Fu, Jiang, Hayashi, and Neubig}]{10.1145/3560815}
Pengfei Liu, Weizhe Yuan, Jinlan Fu, Zhengbao Jiang, Hiroaki Hayashi, and Graham Neubig. 2023{\natexlab{b}}.
\newblock \href {https://doi.org/10.1145/3560815} {Pre-train, prompt, and predict: A systematic survey of prompting methods in natural language processing}.
\newblock \emph{ACM Comput. Surv.}, 55(9).

\bibitem[{Ma et~al.(2023)Ma, Fang, and Wang}]{NEURIPS2023_44956951}
Xinyin Ma, Gongfan Fang, and Xinchao Wang. 2023.
\newblock \href {https://proceedings.neurips.cc/paper_files/paper/2023/file/44956951349095f74492a5471128a7e0-Paper-Conference.pdf} {Llm-pruner: On the structural pruning of large language models}.
\newblock In \emph{Advances in Neural Information Processing Systems}, volume~36, pages 21702--21720. Curran Associates, Inc.

\bibitem[{Mansilla(2004)}]{mansilla2004foundations}
Paloma~Ubeda Mansilla. 2004.
\newblock Foundations of language (brain, meaning, grammar, evolution), by ray jackendoff.
\newblock \emph{Ib{\'e}rica: Revista de la Asociaci{\'o}n Europea de Lenguas para Fines Espec{\'\i}ficos (AELFE)}, (7):150--152.

\bibitem[{Meng et~al.(2024)Meng, Liu, Joty, Xiong, Zhou, and Yavuz}]{SFRAIResearch2024}
Rui Meng, Ye~Liu, Shafiq~Rayhan Joty, Caiming Xiong, Yingbo Zhou, and Semih Yavuz. 2024.
\newblock \href {https://blog.salesforceairesearch.com/sfr-embedded-mistral/} {Sfr-embedding-mistral:enhance text retrieval with transfer learning}.
\newblock Salesforce AI Research Blog.

\bibitem[{Mu et~al.(2024)Mu, Li, and Goodman}]{mu2024learning}
Jesse Mu, Xiang~Lisa Li, and Noah Goodman. 2024.
\newblock Learning to compress prompts with gist tokens.
\newblock In \emph{Proceedings of the 37th International Conference on Neural Information Processing Systems}, NIPS '23, Red Hook, NY, USA. Curran Associates Inc.

\bibitem[{Pan et~al.(2024)Pan, Wu, Jiang, Xia, Luo, Zhang, Lin, R{\"u}hle, Yang, Lin, Zhao, Qiu, and Zhang}]{pan-etal-2024-llmlingua}
Zhuoshi Pan, Qianhui Wu, Huiqiang Jiang, Menglin Xia, Xufang Luo, Jue Zhang, Qingwei Lin, Victor R{\"u}hle, Yuqing Yang, Chin-Yew Lin, H.~Vicky Zhao, Lili Qiu, and Dongmei Zhang. 2024.
\newblock \href {https://aclanthology.org/2024.findings-acl.57} {{LLML}ingua-2: Data distillation for efficient and faithful task-agnostic prompt compression}.
\newblock In \emph{Findings of the Association for Computational Linguistics ACL 2024}, pages 963--981, Bangkok, Thailand and virtual meeting. Association for Computational Linguistics.

\bibitem[{Qiao et~al.(2023)Qiao, Ou, Zhang, Chen, Yao, Deng, Tan, Huang, and Chen}]{qiao-etal-2023-reasoning}
Shuofei Qiao, Yixin Ou, Ningyu Zhang, Xiang Chen, Yunzhi Yao, Shumin Deng, Chuanqi Tan, Fei Huang, and Huajun Chen. 2023.
\newblock \href {https://doi.org/10.18653/v1/2023.acl-long.294} {Reasoning with language model prompting: A survey}.
\newblock In \emph{Proceedings of the 61st Annual Meeting of the Association for Computational Linguistics (Volume 1: Long Papers)}, pages 5368--5393, Toronto, Canada. Association for Computational Linguistics.

\bibitem[{Radford et~al.(2021)Radford, Kim, Hallacy, Ramesh, Goh, Agarwal, Sastry, Askell, Mishkin, Clark, Krueger, and Sutskever}]{pmlr-v139-radford21a}
Alec Radford, Jong~Wook Kim, Chris Hallacy, Aditya Ramesh, Gabriel Goh, Sandhini Agarwal, Girish Sastry, Amanda Askell, Pamela Mishkin, Jack Clark, Gretchen Krueger, and Ilya Sutskever. 2021.
\newblock \href {https://proceedings.mlr.press/v139/radford21a.html} {Learning transferable visual models from natural language supervision}.
\newblock In \emph{Proceedings of the 38th International Conference on Machine Learning}, volume 139 of \emph{Proceedings of Machine Learning Research}, pages 8748--8763. PMLR.

\bibitem[{Radford et~al.(2019)Radford, Wu, Child, Luan, Amodei, Sutskever et~al.}]{radford2019language}
Alec Radford, Jeffrey Wu, Rewon Child, David Luan, Dario Amodei, Ilya Sutskever, et~al. 2019.
\newblock Language models are unsupervised multitask learners.
\newblock \emph{OpenAI blog}, 1(8):9.

\bibitem[{Rajpurkar et~al.(2016)Rajpurkar, Zhang, Lopyrev, and Liang}]{rajpurkar-etal-2016-squad}
Pranav Rajpurkar, Jian Zhang, Konstantin Lopyrev, and Percy Liang. 2016.
\newblock \href {https://doi.org/10.18653/v1/D16-1264} {{SQ}u{AD}: 100,000+ questions for machine comprehension of text}.
\newblock In \emph{Proceedings of the 2016 Conference on Empirical Methods in Natural Language Processing}, pages 2383--2392, Austin, Texas. Association for Computational Linguistics.

\bibitem[{Rau et~al.(2024)Rau, Wang, Déjean, and Clinchant}]{rau2024cocom}
David Rau, Shuai Wang, Hervé Déjean, and Stéphane Clinchant. 2024.
\newblock \href {https://arxiv.org/abs/2407.09252} {Context embeddings for efficient answer generation in rag}.
\newblock \emph{Preprint}, arXiv:2407.09252.

\bibitem[{Ren et~al.(2021)Ren, Dai, Dai, Yang, Leskovec, Schuurmans, and Dai}]{NEURIPS2021_bd4a6d05}
Hongyu Ren, Hanjun Dai, Zihang Dai, Mengjiao Yang, Jure Leskovec, Dale Schuurmans, and Bo~Dai. 2021.
\newblock \href {https://proceedings.neurips.cc/paper_files/paper/2021/file/bd4a6d0563e0604510989eb8f9ff71f5-Paper.pdf} {Combiner: Full attention transformer with sparse computation cost}.
\newblock In \emph{Advances in Neural Information Processing Systems}, volume~34, pages 22470--22482. Curran Associates, Inc.

\bibitem[{Reynolds and McDonell(2021)}]{10.1145/3411763.3451760}
Laria Reynolds and Kyle McDonell. 2021.
\newblock \href {https://doi.org/10.1145/3411763.3451760} {Prompt programming for large language models: Beyond the few-shot paradigm}.
\newblock In \emph{Extended Abstracts of the 2021 CHI Conference on Human Factors in Computing Systems}, CHI EA '21, New York, NY, USA. Association for Computing Machinery.

\bibitem[{Schulhoff et~al.(2024)Schulhoff, Ilie, Balepur, Kahadze, Liu, Si, Li, Gupta, Han, Schulhoff, Dulepet, Vidyadhara, Ki, Agrawal, Pham, Kroiz, Li, Tao, Srivastava, Costa, Gupta, Rogers, Goncearenco, Sarli, Galynker, Peskoff, Carpuat, White, Anadkat, Hoyle, and Resnik}]{schulhoff2024promptreportsystematicsurvey}
Sander Schulhoff, Michael Ilie, Nishant Balepur, Konstantine Kahadze, Amanda Liu, Chenglei Si, Yinheng Li, Aayush Gupta, HyoJung Han, Sevien Schulhoff, Pranav~Sandeep Dulepet, Saurav Vidyadhara, Dayeon Ki, Sweta Agrawal, Chau Pham, Gerson Kroiz, Feileen Li, Hudson Tao, Ashay Srivastava, Hevander~Da Costa, Saloni Gupta, Megan~L. Rogers, Inna Goncearenco, Giuseppe Sarli, Igor Galynker, Denis Peskoff, Marine Carpuat, Jules White, Shyamal Anadkat, Alexander Hoyle, and Philip Resnik. 2024.
\newblock \href {https://arxiv.org/abs/2406.06608} {The prompt report: A systematic survey of prompting techniques}.
\newblock \emph{Preprint}, arXiv:2406.06608.

\bibitem[{Sennrich et~al.(2016)Sennrich, Haddow, and Birch}]{sennrich-etal-2016-neural}
Rico Sennrich, Barry Haddow, and Alexandra Birch. 2016.
\newblock \href {https://doi.org/10.18653/v1/P16-1162} {Neural machine translation of rare words with subword units}.
\newblock In \emph{Proceedings of the 54th Annual Meeting of the Association for Computational Linguistics (Volume 1: Long Papers)}, pages 1715--1725, Berlin, Germany. Association for Computational Linguistics.

\bibitem[{Shandilya et~al.(2024)Shandilya, Xia, Ghosh, Jiang, Zhang, Wu, and Rühle}]{shandilya2024TACO-RL}
Shivam Shandilya, Menglin Xia, Supriyo Ghosh, Huiqiang Jiang, Jue Zhang, Qianhui Wu, and Victor Rühle. 2024.
\newblock \href {https://arxiv.org/abs/2409.13035} {Taco-rl: Task aware prompt compression optimization with reinforcement learning}.
\newblock \emph{Preprint}, arXiv:2409.13035.

\bibitem[{Shen et~al.(2024)Shen, Tenenholtz, Hall, Alvarez-Melis, and Fusi}]{shen2024tag}
Junhong Shen, Neil Tenenholtz, James~Brian Hall, David Alvarez-Melis, and Nicolo Fusi. 2024.
\newblock Tag-llm: Repurposing general-purpose llms for specialized domains.
\newblock \emph{arXiv preprint arXiv:2402.05140}.

\bibitem[{Shin et~al.(2020)Shin, Razeghi, Logan~IV, Wallace, and Singh}]{shin-etal-2020-autoprompt}
Taylor Shin, Yasaman Razeghi, Robert~L. Logan~IV, Eric Wallace, and Sameer Singh. 2020.
\newblock \href {https://doi.org/10.18653/v1/2020.emnlp-main.346} {{A}uto{P}rompt: {E}liciting {K}nowledge from {L}anguage {M}odels with {A}utomatically {G}enerated {P}rompts}.
\newblock In \emph{Proceedings of the 2020 Conference on Empirical Methods in Natural Language Processing (EMNLP)}, pages 4222--4235, Online. Association for Computational Linguistics.

\bibitem[{Tan et~al.(2024)Tan, Li, Patil, Wu, Zhang, Keutzer, Gonzalez, and Popa}]{tan2024LLoCO}
Sijun Tan, Xiuyu Li, Shishir Patil, Ziyang Wu, Tianjun Zhang, Kurt Keutzer, Joseph~E. Gonzalez, and Raluca~Ada Popa. 2024.
\newblock \href {https://arxiv.org/abs/2404.07979} {Lloco: Learning long contexts offline}.
\newblock \emph{Preprint}, arXiv:2404.07979.

\bibitem[{Taori et~al.(2023)Taori, Gulrajani, Zhang, Dubois, Li, Guestrin, Liang, and Hashimoto}]{taori2023alpaca}
Rohan Taori, Ishaan Gulrajani, Tianyi Zhang, Yann Dubois, Xuechen Li, Carlos Guestrin, Percy Liang, and Tatsunori~B Hashimoto. 2023.
\newblock Alpaca: A strong, replicable instruction-following model.
\newblock \emph{Stanford Center for Research on Foundation Models. https://crfm. stanford. edu/2023/03/13/alpaca. html}, 3(6):7.

\bibitem[{Tay et~al.(2022)Tay, Dehghani, Bahri, and Metzler}]{10.1145/3530811}
Yi~Tay, Mostafa Dehghani, Dara Bahri, and Donald Metzler. 2022.
\newblock \href {https://doi.org/10.1145/3530811} {Efficient transformers: A survey}.
\newblock \emph{ACM Comput. Surv.}, 55(6).

\bibitem[{Teehan et~al.(2024)Teehan, Lake, and Ren}]{teehan2024college}
Ryan Teehan, Brenden Lake, and Mengye Ren. 2024.
\newblock \href {https://openreview.net/forum?id=Fkr1yVUb9G} {Co{LLEG}e: Concept embedding generation for large language models}.
\newblock In \emph{First Conference on Language Modeling}.

\bibitem[{Todd et~al.(2024)Todd, Li, Sharma, Mueller, Wallace, and Bau}]{todd2023function}
Eric Todd, Millicent Li, Arnab~Sen Sharma, Aaron Mueller, Byron~C Wallace, and David Bau. 2024.
\newblock \href {https://openreview.net/forum?id=AwyxtyMwaG} {Function vectors in large language models}.
\newblock In \emph{The Twelfth International Conference on Learning Representations}.

\bibitem[{Touvron et~al.(2023)Touvron, Martin, Stone, Albert, Almahairi, Babaei, Bashlykov, Batra, Bhargava, Bhosale et~al.}]{touvron2023llama}
Hugo Touvron, Louis Martin, Kevin Stone, Peter Albert, Amjad Almahairi, Yasmine Babaei, Nikolay Bashlykov, Soumya Batra, Prajjwal Bhargava, Shruti Bhosale, et~al. 2023.
\newblock Llama 2: Open foundation and fine-tuned chat models.
\newblock \emph{arXiv preprint arXiv:2307.09288}.

\bibitem[{Vatsal and Dubey(2024)}]{vatsal2024surveypromptengineeringmethods}
Shubham Vatsal and Harsh Dubey. 2024.
\newblock \href {https://arxiv.org/abs/2407.12994} {A survey of prompt engineering methods in large language models for different nlp tasks}.
\newblock \emph{Preprint}, arXiv:2407.12994.

\bibitem[{Wan et~al.(2024)Wan, Wang, Liu, Alam, Zheng, Liu, Qu, Yan, Zhu, Zhang, Chowdhury, and Zhang}]{wan2024efficient}
Zhongwei Wan, Xin Wang, Che Liu, Samiul Alam, Yu~Zheng, Jiachen Liu, Zhongnan Qu, Shen Yan, Yi~Zhu, Quanlu Zhang, Mosharaf Chowdhury, and Mi~Zhang. 2024.
\newblock \href {https://openreview.net/forum?id=bsCCJHbO8A} {Efficient large language models: A survey}.
\newblock \emph{Transactions on Machine Learning Research}.
\newblock Survey Certification.

\bibitem[{Wingate et~al.(2022)Wingate, Shoeybi, and Sorensen}]{wingate-etal-2022-prompt}
David Wingate, Mohammad Shoeybi, and Taylor Sorensen. 2022.
\newblock \href {https://doi.org/10.18653/v1/2022.findings-emnlp.412} {Prompt compression and contrastive conditioning for controllability and toxicity reduction in language models}.
\newblock In \emph{Findings of the Association for Computational Linguistics: EMNLP 2022}, pages 5621--5634, Abu Dhabi, United Arab Emirates. Association for Computational Linguistics.

\bibitem[{Xu et~al.(2024{\natexlab{a}})Xu, Shi, and Choi}]{xu2024recomp}
Fangyuan Xu, Weijia Shi, and Eunsol Choi. 2024{\natexlab{a}}.
\newblock \href {https://openreview.net/forum?id=mlJLVigNHp} {{RECOMP}: Improving retrieval-augmented {LM}s with context compression and selective augmentation}.
\newblock In \emph{The Twelfth International Conference on Learning Representations}.

\bibitem[{Xu et~al.(2024{\natexlab{b}})Xu, Feng, Mu, Hou, Li, Wang, Zhong, Li, Tu, Zhu, Zhang, and Che}]{xu2024conciseprecisecontextcompression}
Yang Xu, Yunlong Feng, Honglin Mu, Yutai Hou, Yitong Li, Xinghao Wang, Wanjun Zhong, Zhongyang Li, Dandan Tu, Qingfu Zhu, Min Zhang, and Wanxiang Che. 2024{\natexlab{b}}.
\newblock \href {https://arxiv.org/abs/2407.02043} {Concise and precise context compression for tool-using language models}.
\newblock \emph{Preprint}, arXiv:2407.02043.

\bibitem[{Yang et~al.(2018)Yang, Qi, Zhang, Bengio, Cohen, Salakhutdinov, and Manning}]{yang-etal-2018-hotpotqa}
Zhilin Yang, Peng Qi, Saizheng Zhang, Yoshua Bengio, William Cohen, Ruslan Salakhutdinov, and Christopher~D. Manning. 2018.
\newblock \href {https://doi.org/10.18653/v1/D18-1259} {{H}otpot{QA}: A dataset for diverse, explainable multi-hop question answering}.
\newblock In \emph{Proceedings of the 2018 Conference on Empirical Methods in Natural Language Processing}, pages 2369--2380, Brussels, Belgium. Association for Computational Linguistics.

\bibitem[{yang Liu et~al.(2024)yang Liu, Wang, Yin, Molchanov, Wang, Cheng, and Chen}]{liu2024dora}
Shih yang Liu, Chien-Yi Wang, Hongxu Yin, Pavlo Molchanov, Yu-Chiang~Frank Wang, Kwang-Ting Cheng, and Min-Hung Chen. 2024.
\newblock \href {https://openreview.net/forum?id=3d5CIRG1n2} {Do{RA}: Weight-decomposed low-rank adaptation}.
\newblock In \emph{Forty-first International Conference on Machine Learning}.

\bibitem[{Yoon et~al.(2024)Yoon, Lee, Hwang, Jeong, and Kang}]{yoon2024compact}
Chanwoong Yoon, Taewhoo Lee, Hyeon Hwang, Minbyul Jeong, and Jaewoo Kang. 2024.
\newblock \href {https://arxiv.org/abs/2407.09014} {Compact: Compressing retrieved documents actively for question answering}.
\newblock \emph{Preprint}, arXiv:2407.09014.

\bibitem[{Zhang et~al.(2024{\natexlab{a}})Zhang, Huang, Jin, and Lu}]{10445007}
Jingyi Zhang, Jiaxing Huang, Sheng Jin, and Shijian Lu. 2024{\natexlab{a}}.
\newblock \href {https://doi.org/10.1109/TPAMI.2024.3369699} {Vision-language models for vision tasks: A survey}.
\newblock \emph{IEEE Transactions on Pattern Analysis and Machine Intelligence}, 46(8):5625--5644.

\bibitem[{Zhang et~al.(2024{\natexlab{b}})Zhang, Zhang, Pang, Zheng, and Zheng}]{zhang2024adacomp}
Qianchi Zhang, Hainan Zhang, Liang Pang, Hongwei Zheng, and Zhiming Zheng. 2024{\natexlab{b}}.
\newblock \href {https://arxiv.org/abs/2409.01579} {Adacomp: Extractive context compression with adaptive predictor for retrieval-augmented large language models}.
\newblock \emph{Preprint}, arXiv:2409.01579.

\bibitem[{Zhang et~al.(2024{\natexlab{c}})Zhang, Han, Liu, Zhou, Lu, Qiao, Li, and Gao}]{zhang2024llamaadapter}
Renrui Zhang, Jiaming Han, Chris Liu, Aojun Zhou, Pan Lu, Yu~Qiao, Hongsheng Li, and Peng Gao. 2024{\natexlab{c}}.
\newblock \href {https://openreview.net/forum?id=d4UiXAHN2W} {{LL}a{MA}-adapter: Efficient fine-tuning of large language models with zero-initialized attention}.
\newblock In \emph{The Twelfth International Conference on Learning Representations}.

\bibitem[{Zhao et~al.(2023)Zhao, Gupta, Chung, and Huang}]{zhao-etal-2023-spc}
Wenbo Zhao, Arpit Gupta, Tagyoung Chung, and Jing Huang. 2023.
\newblock \href {https://doi.org/10.18653/v1/2023.repl4nlp-1.10} {{SPC}: Soft prompt construction for cross domain generalization}.
\newblock In \emph{Proceedings of the 8th Workshop on Representation Learning for NLP (RepL4NLP 2023)}, pages 118--130, Toronto, Canada. Association for Computational Linguistics.

\end{thebibliography}

\appendix

\section{Appendix}

\begin{figure*}[th]
    \centering
    \includegraphics[width=0.95\textwidth]{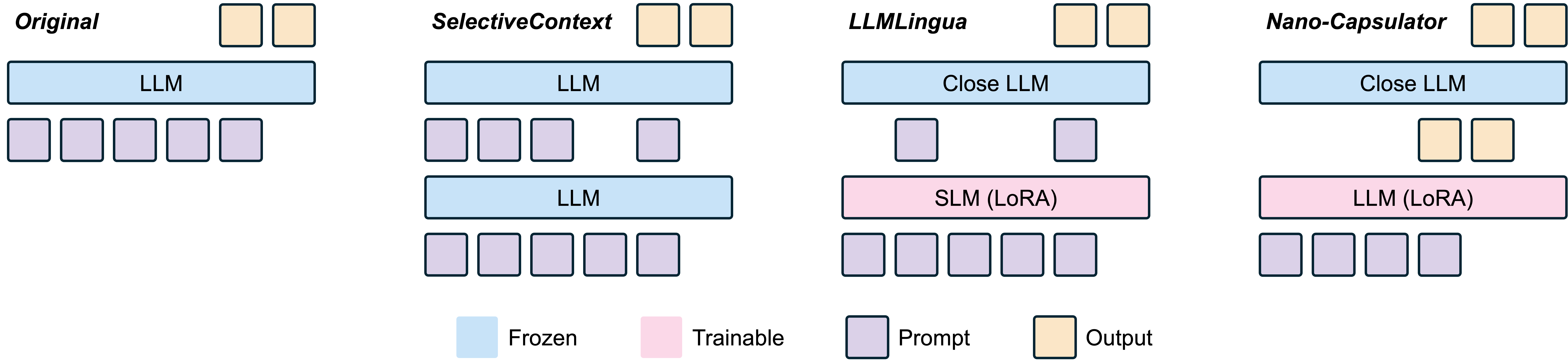}
    \caption{Architectures for various prompt compression models by hard prompt methods.}
    \label{hard_prompt_methods}
\end{figure*}

\begin{figure*}[th!]
    \centering
    \includegraphics[width=0.95\textwidth]{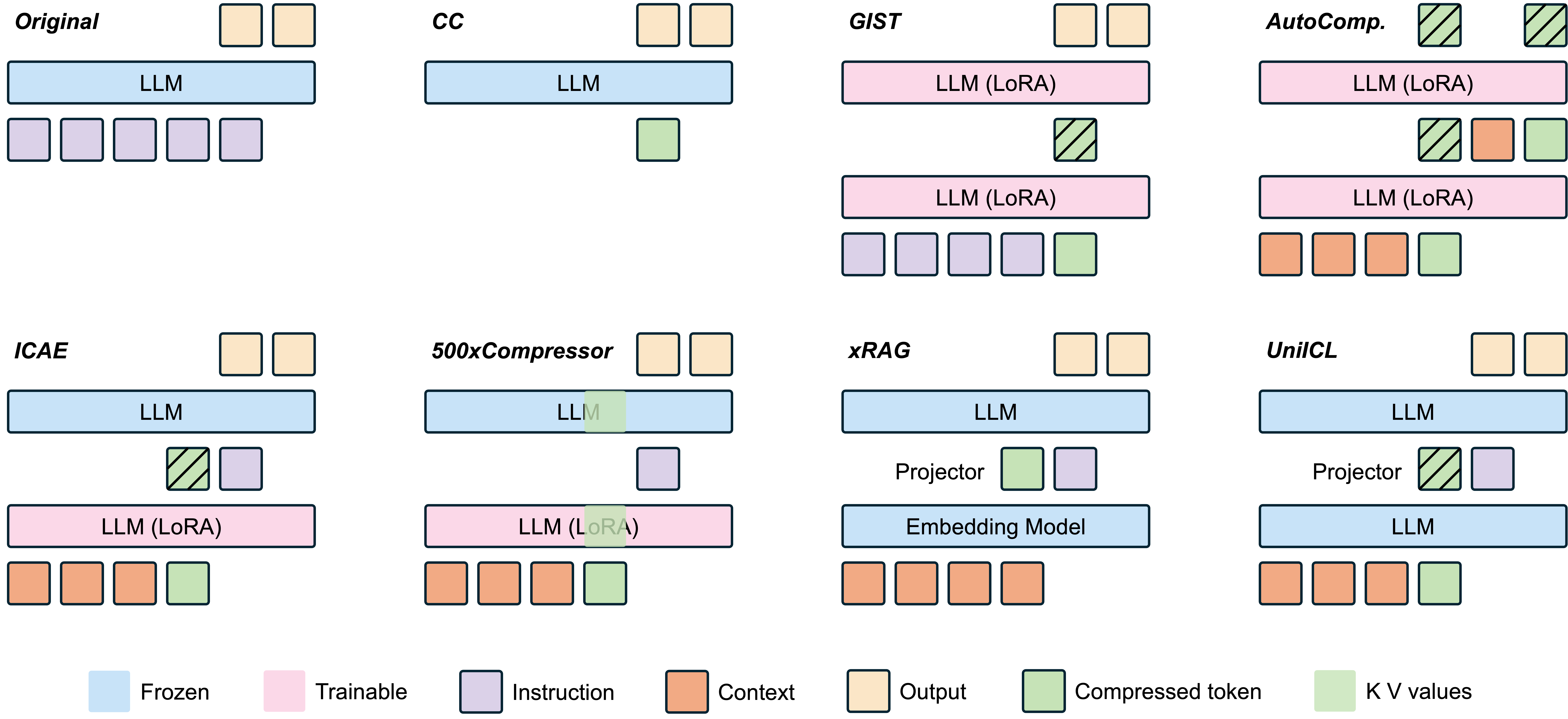}
    \caption{Architectures for various prompt compression models by soft prompt methods.}
    \label{soft_prompt_methods}
    \vspace{-4mm}
\end{figure*}

To provide readers with an interactive way to locate reference papers on prompt compression methods, Figure~\ref{tree_overview} has been translated into Table~\ref{overview_table}. Data examples for prompt structures in Section~\ref{Preliminary} are presented in Table~\ref{prompt_structures}. Figures~\ref{hard_prompt_methods} and \ref{soft_prompt_methods} compare the architectures of various hard and soft prompt methods for prompt compression.

This paper was refined with ChatGPT.

\begin{table*}[th!]
    \centering
    \begin{tabular}{p{0.95\linewidth}}
        \hline
        \textbf{Hard Prompt Methods} \\
        \textbf{\textit{Filtering:}} \\
        General: SelectiveContext \citep{li-etal-2023-compressing}, LLMLingua \citep{jiang-etal-2023-llmlingua}, LongLLMLingua \citep{jiang-etal-2024-longllmlingua}, AdaComp \citep{zhang2024adacomp} \\
        Distillation Enhanced: LLMLingua-2 \citep{pan-etal-2024-llmlingua} \\
        RL Enhanced: TACO-RL \citep{shandilya2024TACO-RL}, PCRL \citep{Jung_2024} \\
        Embedding Enhanced: CPC \citep{liskavets2024CPC}, TCRA-LLM \citep{liu-etal-2023-tcra} \\
        \textbf{\textit{Paraphrasing:}} \\
        Nano-Capsulator \citep{chuang-etal-2024-learning}, CompAct \citep{yoon2024compact}, FAVICOMP \citep{jung2024FaviComp} \\ 
        \hline
        \textbf{Soft Prompt Methods} \\ 
        \textbf{\textit{Decoder Only:}} \\
        Not Finetuned: CC \citep{wingate-etal-2022-prompt} \\
        Finetuned: GIST \citep{mu2024learning}, AutoCompressor \citep{chevalier2023adapting} \\
        \textbf{\textit{Encoder-decoder:}} \\
        Both Finetuned: COCOM \citep{rau2024cocom}, LLoCO \citep{tan2024LLoCO} \\
        Finetuned Encoder: ICAE \citep{ge2024incontext}, 500xCompressor \citep{500xcompressor}, QGC \citep{cao-etal-2024-retaining} \\
        Embedding Encoder: xRAG \citep{cheng2024xragextremecontextcompression} \\
        Projector: UniICL \citep{gao2024unifyingdemonstrationselectioncompression} \\ 
        \hline
        \textbf{Applications:} \\
        \textbf{\textit{RAG:}} \\
        RAG: xRAG \citep{cheng2024xragextremecontextcompression}, RECOMP \citep{xu2024recomp}, COCOM \citep{rau2024cocom}, CompAct \citep{yoon2024compact}, FAVICOMP \citep{jung2024FaviComp}, AdaComp \citep{zhang2024adacomp}, LLoCO \citep{tan2024LLoCO}, TCRA-LLM \citep{liu-etal-2023-tcra} \\
        Agents: HD-Gist \citep{jiang-etal-2024-hierarchical}, \citep{xu2024conciseprecisecontextcompression} \\
        Domain-specific tasks: Tag-llm \citep{shen2024tag}, CoLLEGe \citep{teehan2024college} \\
        Others: ICL \citep{gao2024unifyingdemonstrationselectioncompression}, Role Playing \citep{NEURIPS2023_6fcbfb37}, Functions \citep{todd2023function} \\
        \hline
    \end{tabular}
    \caption{Summary of methods and applications in prompt compression.}
    \label{overview_table}
\end{table*}

\begin{table*}[th!]
    \centering
    \begin{tabular}{p{0.95\linewidth}}
        \hline
        \textbf{\{Instruction, Input, Output\}} \\
        \textbf{\textit{Example in Alpaca:}} \\ 
        Instruction: Evaluate this sentence for spelling and grammar mistakes\\
        Input: He finnished his meal and left the resturant\\
        Output: There are two spelling errors in the sentence. The corrected sentence should be: "He finished his meal and left the restaurant."\\
        \hline
        \textbf{\{Context, Question, Answer\}} \\ 
        \textbf{\textit{Example in SQuAD:}} \\
        Context: Super Bowl 50 was an Americould football game to determine the champion of the National Football League (NFL) for the 2015 season. The Americould Football Conference (AFC) champion Denver Broncos defeated the National Football Conference (NFC) champion Carolina Panthers 24–10 to earn their third Super Bowl title. The game was played on February 7, 2016, at Levi's Stadium in the San Francisco Bay Area at Santa Clara, California. As this was the 50th Super Bowl, the league emphasized the "golden anniversary" with various gold-themed initiatives, as well as temporarily suspending the tradition of naming each Super Bowl game with Roman numerals (under which the game would have been known as "Super Bowl L"), so that the logo can prominently feature the Arabic numerals 50. \\
        Question: Which NFL team represented the AFC at Super Bowl 50? \\
        Answer: Denver Broncos \\
        \hline
    \end{tabular}
    \caption{Examples for the QA items with formats of \{Instruction, Input, Output\} and \{Context, Question, Answer\}.}
    \label{prompt_structures}
\end{table*}

\end{document}